\documentclass{article}
\usepackage{spconf}
\usepackage{cite}
\usepackage{color}
\usepackage[pdftex]{graphicx}
\graphicspath{{fig/}{jpeg/}}
\usepackage[cmex10]{amsmath}
\usepackage{amssymb}
\usepackage{algorithm}
\usepackage{algpseudocode}
\usepackage{comment}
\input{mysymbol.sty}
\usepackage{needspace}

\usepackage{theorem}
\newtheorem{thm}{Theorem}
\newtheorem{lemma}{Lemma}
\theoremstyle{definition}
\newtheorem{assumption}{Assumption}

\ninept

\name{Aryan Mokhtari and Alejandro Ribeiro \thanks{Supported by NSF CAREER CCF-0952867 and ONR N00014-12-1-0997.}}
\address{Department of Electrical and Systems Engineering, University of Pennsylvania}

\title{\vspace{-10mm} A Quasi-Newton method for large scale support vector machines}

\begin{document}

\maketitle

\pagestyle{empty}

\begin{abstract}
This paper adapts a recently developed regularized stochastic version of the Broyden, Fletcher, Goldfarb, and Shanno (BFGS) quasi-Newton method for the solution of support vector machine classification problems. The proposed method is shown to converge almost surely to the optimal classifier at a rate that is linear in expectation. Numerical results show that the proposed method exhibits a convergence rate that degrades smoothly with the dimensionality of the feature vectors.

\end{abstract}

\section{Introduction}
\label{sec:Intro}

Given a training set with points whose class is known the goal of a support vector machine (SVM) is to find a hyperplane that best separates the training set. If future samples are statistically identical to the training set this hyperplane provides the best classification accuracy. Computation of the separating hyperplane entails solution of a convex optimization problem that can be implemented without much difficulty in problems of moderate size \cite{Bottou}. Large scale problems in which the dimension of the points to be classified is large require a commensurably large training set. In these situations, computing the gradients that are required for numerical determination of the separating hyperplanes becomes infeasible and motivates the use of stochastic gradient descent methods which build unbiased gradient estimates based on small data subsamples \cite{Bottou,Shwartz,Zhang,Roux}.

However practical, stochastic gradient descent methods need a large number of iterations to converge. This translates into the need of very large training sets, or, since the size of the training set is in general limited by data collection, in the computation of hyperplanes that are not as good classifiers as they could be given the available data. In this paper we resort to quasi-Newton methods \cite{Broyden, Bordes, Byrd, AryanAle, cMokhtariRibeiro13, Nocedal, Powell, Schraudolph} to make better use of the provided training set. In particular, we adapt a recently developed regularized stochastic version of the Broyden, Fletcher, Goldfarb, and Shanno (BFGS) method\cite{cMokhtariRibeiro13} for the solution of SVM classification problems (Section \ref{sec:problem}). The proposed method is shown to converge almost surely over realizations of the training set to the optimal classifier (Theorem \ref{convg}) at a rate that is linear in expectation (Theorem  \ref{theo_convergence_rate}). Numerical results show that the method exhibits a convergence time that degrades smoothly with the dimensionality of the feature vectors. (Section \ref{sec:SVM}).

\vspace{-1mm}
\section{Stochastic quasi-Newton method} \label{sec:problem}

Consider a training set $\ccalS = \{ (\bbx_{i},y_{i}) \}_{i=1}^{N}$ containing $N$ pairs of the form $(\bbx_{i},y_i)$, where $\bbx_{i}\in\reals^n$ is a feature vector and $y_{i}\in \{-1,1 \}$ the corresponding vector's class. We want to find a hyperplane supported by a vector $\bbw\in\reals^n$ which separates the training set so that $\bbw^{T}\bbx_i>0$ for all points with $y_i=1$ and $\bbw^{T}\bbx_i<0$ for all points with $y_i=-1$. Since this vector may not exist if the data is not perfectly separable we introduce the loss function $l((\bbx,y);\bbw)$ measuring the distance between the point $\bbx_i$ and the hyperplane supported by $\bbw$ and proceed to select the hyperplane supporting vector as the one with minimum aggregate loss
\begin{equation}\label{SVM}
   \bbw^* := \argmin\   \frac{\lambda}{2}\|\bbw\|^2 
                      + \frac{1}{N} \sum_{i=1}^{N} l((\bbx_{i},y_{i});\bbw),
\end{equation}
where we also added the regularization term $\lambda\|\bbw\|^2/2 $ for some constant $\lambda>0$. The vector $\bbw^*$ in \eqref{SVM} balances the minimization of the sum of distances to the separating hyperplane, as measured by the loss function $l((\bbx,y);\bbw)$, with the minimization of the $L_{2}$ norm $\|\bbw\|_2$ to enforce desirable properties in $\bbw^*$ {\cite{Vapnik}}. 
Common selections for the loss function are the squared hinge loss $l((\bbx,y);\bbw)=\max(0,1-y(\bbw^{T}\bbx))^{2}$ and the log loss $l((\bbx,y);\bbv)=\log(1+\exp(-y(\bbw^{T}\bbx)))$, e.g.{\cite{Bottou}}. 

To model \eqref{SVM} as a stochastic optimization problem let $\bbtheta_i:=(\bbx_{i},y_i)$ be a given training point and consider a uniform probability distribution on the training set $\ccalS = \{ (\bbx_{i},y_{i}) \}_{i=1}^{N} =\{ \bbtheta_{i} \}_{i=1}^{N}$. Upon defining the function $f(\bbw,\bbtheta) :=\lambda\|\bbw\|^2/2 + l((\bbx_{i},y_{i});\bbw)$ we can rewrite  \eqref{SVM} as
\begin{equation}\label{optimization_problem}
   \bbw^* := \argmin_\bbw \mbE_{\bbtheta}[f(\bbw,{\bbtheta})]
          := \argmin_\bbw {F(\bbw)}.
\end{equation} 
In \eqref{optimization_problem}, we (re-)interpret the sum in \eqref{SVM} as an expectation over the uniform discrete distribution on the set $\ccalS$. We refer to $f(\bbw,{\bbtheta})$ as the instantaneous functions and to $F(\bbw):=\mbE_{\bbtheta}[f(\bbw,{\bbtheta})]$ as the average function. 

Since the loss functions $l((\bbx_{i},y_{i});\bbw)$ are convex, the functions $f(\bbw,\bbtheta) :=\lambda\|\bbw\|^2/2 + l((\bbx_{i},y_{i});\bbw)$ are strongly convex. Thus, the average objective $F(\bbw)$ in \eqref{optimization_problem} is also strongly convex and the optimal separating hyperplane $\bbw^{*}$ can be found by stochastic gradient descent algorithms. However, the number of iterations required to run these algorithms, which translates to the number of training features $(\bbx_{i},y_{i})$ that need to be acquired, becomes prohibitive for large dimensional problems. To reduce the number of iterations required for convergence we develop a regularized stochastic version of the BFGS method.

To be precise let $t\geq0$ be an iteration index and assume that at time $t$ we are given a sample of $L$ realizations of the random variables $\bbtheta$. Group these samples in the vector $\tbtheta_t := [\bbtheta_{t1};...;\bbtheta_{tL}]$ and let $\bbw_t$ denote the current hyperplane normal vector iterate. We then define the stochastic gradient of $F(\bbw)$ associated with samples $\tbtheta_t$ at point $\bbw_t$ as
\begin{equation}\label{stochastic_gradient}
   \hbs(\bbw_t,\tbtheta_t) 
          = \frac{1}{L}\sum_{l=1}^{L}  \nabla f(\bbw_t,{\bbtheta_{tl}}).
\end{equation}
Further introduce a step size sequence $\eps_t$, a positive definite curvature approximation matrix $\hbB_{t}$, and a regularization constant $\Gamma>0$. The regularized stochastic BFGS algorithm is then defined by the iteration
\begin{equation}\label{eqn_sbfgs_dual_iteration}
   \bbw_{t+1} = \bbw_{t}-\epsilon_{t}\ \left(\hbB_{t}^{-1}+\Gamma \bbI\right)\ \hbs(\bbw_{t},\tbtheta_{t}).
\end{equation}
The update in \eqref{eqn_sbfgs_dual_iteration} proceeds along the negative stochastic gradient direction $-\hbs(\bbw_{t},\tbtheta_{t})$ premultiplied by the positive definite matrix $\hbB_{t}^{-1}+\Gamma \bbI$ and modulated by the step size $\eps_t$. 

For the algorithm in \eqref{eqn_sbfgs_dual_iteration} to have better convergence properties than gradient descent we need the matrix $\hbB_{t}$ to approximate  the Hessian of the objective function $\bbH(\bbw_t):=\nabla^2 F(\bbw_t)$ so that \eqref{eqn_sbfgs_dual_iteration} approximates an stochastic version of Newton's method -- the role of $\Gamma \bbI$ is to provide a guarantee of minimum progress as we discuss in the convergence analysis in Section \ref{sec_convergence}. To define such approximation we use a stochastic version of the secant condition used in deterministic BFGS. Start by defining the variable and stochastic gradient variations at time $t$ as
\begin{equation}\label{chris}
   \bbv_{t} := \bbw_{t+1}-\bbw_{t}, \qquad
   \hbr_{t} :=\hbs(\bbw_{t+1},\tbtheta_{t})-\hbs(\bbw_{t},\tbtheta_{t}), 
\end{equation}
respectively, and select the matrix $\hbB_{t+1}$ to be used in the next time step so that it satisfies the secant condition $ \hbB_{t+1} \bbv_{t} = \hbr_{t}$. The rationale for this selection is that {the Hessian $\bbH(\bbw_t)$ satisfies this condition for $\bbw_{t+1}$ tending to $\bbw_{t}$}. Notice however that the secant condition $ \hbB_{t+1} \bbv_{t} = \hbr_{t}$ is not enough to completely specify $\hbB_{t+1}$. To resolve this indeterminacy, matrices $\hbB_{t+1}$ in BFGS are also required to be as close as possible to $\hbB_{t}$ in terms of minimizing the Gaussian differential entropy,
\begin{alignat}{2}\label{jadid_prelim}
   \hbB_{t+1} = 
      &\argmin_{\bbZ}\ && \tr\left[\hbB_{t}^{-1}\bbZ\right]
                             -\log\det\left[\hbB_{t}^{-1}\bbZ\right]-n,\nonumber\\
      &\st      && \bbZ \bbv_{t} = \hbr_{t},\quad 
                                     \bbZ \succeq\bbzero.
\end{alignat}
The constraint $\bbZ \succeq\bbzero$ restricts the feasible space to positive semidefinite matrices whereas the constraint $\bbZ \bbv_{t} =  \hbr_{t}$ requires $\bbZ$ to satisfy the secant condition. The objective $\tr(\hbB_{t}^{-1}\bbZ)-\log\det(\hbB_{t}^{-1}\bbZ)-n$ is the differential entropy between Gaussian variables with covariances $\hbB_t$ and $\bbZ$. 

Observe that $\hbB_{t+1}$ stays positive definite as long as the matrix $\hbB_t\succ\bbzero$ is positive definite, e.g. \cite{Nocedal}. However, it is possible for the smallest eigenvalue of $\hbB_t$ to become arbitrarily close to zero which means that the largest eigenvalue of $\hbB_{t}^{-1}$ becomes very large. To avoid this problem we introduce a regularization of \eqref{jadid_prelim} that requires the smallest eigenvalue of $\hbB_{t+1}$ to be larger than a positive constant $\delta$,
\begin{alignat}{2}\label{jadid}
   \hbB_{t+1} = 
      &\argmin_{\bbZ}\ && \tr\left[\hbB_{t}^{-1}(\bbZ-\delta \bbI)\right]
                             -\log\det\left[\hbB_{t}^{-1}(\bbZ-\delta \bbI)\right]-n,\nonumber\\
      &\st      && \bbZ \bbv_{t} = \hbr_{t},\quad 
                                     \bbZ \succeq\bbzero.
\end{alignat}
Since the logarithm determinant $\log\det[\hbB_{t}^{-1}(\bbZ-\delta \bbI)]$ diverges as the smallest eigenvalue of $\bbZ$ approaches $\delta$, the smallest eigenvalue of the Hessian approximation matrices $\hbB_{t+1}$ computed as solutions of \eqref{jadid} exceeds the lower bound $\delta$. Thus, the largest eigenvalue of $\hbB_{t+1}^{-1}$ is bounded above by $1/\delta$. The following lemma shows that solutions of \eqref{jadid} can be computed by a simple algebraic formula (see \cite{AryanAleTSP} for proofs of results in this paper).

%
\begin{lemma}\label{flen}
Consider the semidefinite program in \eqref{jadid} where the matrix $\hbB_{t}\succ\bbzero$ is positive definite and define the corrected gradient variation
\begin{equation}\label{chris2}
   \tbr_{t} := \hbr_{t} - \delta \bbv_{t},
\end{equation}
If $\tbr_{t}^T\bbv_t=(\hbr_t-\delta\bbv_t)^T\bbv_t>0$, the solution $\hbB_{t+1}$ of \eqref{jadid} can be written as
\begin{equation}\label{akbar}
   \hbB_{t+1} = \hbB_{t} + {{\tbr_{t}\tbr_{t}^{T}}\over{\bbv_{t}^{T}\tbr_{t}}}
- {{\hbB_{t} \bbv_{t}\bbv_{t}^{T}{\hbB_{t}} }\over{\bbv_{t}^{T}\hbB_{t}\bbv_{t}}} +\delta \bbI .
\end{equation}
\end{lemma}

%
When $\delta=0$ the update in \eqref{akbar} coincides with standard non-regularized BFGS \cite{Dennis,Powell,Byrd, Nocedal}. Therefore, the differences between BFGS and regularized BFGS are the replacement of the gradient variation $\hbr_t$ by the corrected variation $\tbr_t:=\hbr_t-\delta\bbv_t$ and the addition of the regularization term $\delta\bbI$. Notice that the expression in \eqref{akbar} is the solution to \eqref{jadid} only when the inner product $\tbr_{t}^T\bbv_t=(\hbr_t-\delta\bbv_t)^T\bbv_t>0$. 

\vspace{-1mm}

%
{\begin{algorithm}[t] \label{algo_stochastic_bfgs} 
\caption{Regularized stochastic BFGS support vector machines}
{\footnotesize \begin{algorithmic}[1]
\Require Variable $\bbw_0$. Hessian approximation  $\hbB_0 \succ\delta \bbI$.
\For {$t= 0,1,2,\ldots$}
\State Collect $L$ training points $\tbx_t=[\bbx_{t1},\ldots,\bbx_{tL}]$ 
       and $\tby_t=[y_{t1},\ldots,y_{tL}]$
\State Compute stochastic gradient 
       $\hbs(\bbw_{t},(\tbx_{t},\tby_{t}))$ [cf. \eqref{stochastic_gradient_SVM}].
       \begin{equation*}
          \hbs(\bbw_{t},(\tbx_{t},\tby_{t})) 
               =  \lambda\bbw_{t} 
               + \frac{1}{L} \sum_{i=1}^{L} \nabla_{\bbw}\l((\bbx_{ti},y_{ti});\bbw_{t}).
       \end{equation*}
\State Descend along direction $(\hbB_{t}^{-1}+\Gamma \bbI)\  \hbs(\bbw_{t},(\tbx_{t},\tby_{t}))$ 
          [cf. \eqref{eqn_sbfgs_dual_iteration}]

          \begin{equation*}
             \bbw_{t+1}=\bbw_{t}-\epsilon_{t}\ (\hbB_{t}^{-1}+\Gamma \bbI)\  \hbs(\bbw_{t},(\tbx_{t},\tby_{t})).
          \end{equation*}
   \State Compute $\hbs(\bbw_{t+1},(\tbx_{t},\tby_{t}))$ [cf. \eqref{stochastic_gradient_SVM}]
\begin{equation*}
   \hbs(\bbw_{t+1},(\tbx_{t},\tby_{t}) )
          = \lambda\bbw_{t+1}   + \frac{1}{L} \sum_{i=1}^{L} \nabla_{\bbw}\ \l((\bbx_{ti},y_{ti});\bbw_{t+1}).
\end{equation*}
   \State Variable and modified stochastic gradient variations 
          [cf. \eqref{chris} and  \eqref{chris2}]
          \begin{align*}    
              &\bbv_{t} = \bbw_{t+1}-\bbw_{t}, \\ 
              &\tbr_{t}=\hbs(\bbw_{t+1},(\tbx_{t},\tby_{t}))
               -\hbs(\bbw_{t},(\tbx_{t},\tby_{t}))-\delta \bbv_{t}
          \end{align*}
   \State Update Hessian approximation matrix
          [cf. \eqref{akbar}]

          \begin{equation*}            
  					 \hbB_{t+1} = \hbB_{t} + 			  		{{\tbr_{t}\tbr_{t}^{T}}\over{\bbv_{t}^{T}\tbr_{t}}}
	- {{\hbB_{t} \bbv_{t}\bbv_{t}^{T}{\hbB_{t}} 	}\over{\bbv_{t}^{T}\hbB_{t}\bbv_{t}}} +\delta \bbI .
          \end{equation*}
\EndFor  
\end{algorithmic}}
\end{algorithm}}

\subsection{Regularized stochastic BFGS support vector machines} 
\label{sec:SVMproblem}

To solve the SVM problem in \eqref{SVM} using regularized stochastic BFGS we need the stochastic gradient in \eqref{stochastic_gradient}. For that, select a sample of $L$ feature vectors $\tbx=[\bbx_{1};...;\bbx_{L} ]$ and corresponding classes $\tby=[y_{1};...;y_{L} ]$ from the training set and compute the stochastic gradient as [cf. \eqref{stochastic_gradient}]
\begin{equation}\label{stochastic_gradient_SVM}
\hbs(\bbw,(\tbx,\tby)) =  \lambda\bbw   + \frac{1}{L} \sum_{i=1}^{L} \nabla_{\bbw}\ \l((\bbx_{i},y_{i});\bbw).
\end{equation}     
Start at time $t$ with current iterate $\bbw_t$ and recall that $\hbB_{t}$ stands for the Hessian approximation computed by stochastic BFGS in the previous iteration. Proceed to collect feature vectors $\tbx_t=[\bbx_{t1};...;\bbx_{tL}]$ and their corresponding class vectors $\tby_{t}=[y_{t1};...;y_{tL} ]$ and for each pair $(\tbx_{t},\tby_{t})$ determine the stochastic gradients $\hbs(\bbw_{t},(\tbx_{t},\tby_{t}))$ as per \eqref{stochastic_gradient_SVM}. Descend then along the direction $(\hbB_{t}^{-1}+\Gamma \bbI)\  \hbs(\bbw_{t},(\tbx_{t},\tby_{t}))$ as per \eqref{eqn_sbfgs_dual_iteration}. This leads to the next iterate $\bbw_{t+1}$, but to complete the iteration we still need to compute the updated Hessian approximation $\hbB_{t+1}$. To do so compute the stochastic gradient $\hbs(\bbw_{t+1},(\tbx_{t},\tby_{t}))$ associated with the {\it same} set of random data points samples $(\tbx_t,\tby_{t})$ used to compute the stochastic gradient $\hbs(\bbw_{t},(\tbx_{t},\tby_{t}))$. The stochastic gradient variation $\hbr_{t}$, the variable variation $\bbv_{t}$, and the modified stochastic gradient variation $\tbr_{t}$ at time $t$ are now computed using \eqref{chris} and \eqref{chris2}. The Hessian approximation $\hbB_{t+1}$ for the next iteration is defined as the matrix that satisfies the stochastic secant condition $\hbB_{t+1} \bbv_{t} =  \hbr_{t}$ and is closest to $\hbB_t$ in the sense of \eqref{jadid}. As per Lemma \ref{flen} we can compute $\hbB_{t+1}$ using \eqref{akbar}.

The solution of \eqref{SVM} using regularized stochastic BFGS is summarized in Algorithm $1$. The two core steps in each iteration are the descent in Step 4 and the update of the Hessian approximation $\hbB_{t}$ in Step 8. Step 2 comprises the observation of $L$ pairs of data points and feature vectors that are required to compute the stochastic gradients in steps 3 and 5. The stochastic gradient $\hbs(\bbw_{t},(\tbx_{t},\tby_{t}))$ in Step 3 is used in the descent iteration in Step 4. The stochastic gradient of Step 3 along with the stochastic gradient $\hbs(\bbw_{t+1},(\tbx_{t},\tby_{t}))$ of Step 5 are used to compute the variations in steps 6 and 7 that permit carrying out the update of the Hessian approximation $\hbB_{t}$ in Step 8. Iterations are initialized with arbitrary vector $\bbw_0$ and matrix $\hbB_{0}$ having all eigenvalues larger than $\delta$.
\vspace{-1mm}

\section{Convergence analysis }\label{sec_convergence}

Our goal here is to show that as time progresses the sequence of classifiers $\bbw_t$ approaches the optimal classifier $\bbw^*$. In proving this result we make the following assumptions.

%
\begin{assumption}\label{ass_intantaneous_hessian_bounds}\normalfont  For any set of samples $\tbtheta=[\bbtheta_1,\ldots,\bbtheta_L]$ the instantaneous functions $\hhatf(\bbw,{\tbtheta}):= (1/L)\sum_{l=1}^L f(\bbw,{\bbtheta_{l}})$ are twice differentiable and their Hessians $\hat{ \bbH}(\bbw,{\tbtheta} )=\nabla_{\bbw}^2\hhatf(\bbw,{\tbtheta})$ have lower and upper bounded eigenvalues,
\begin{equation}\label{hassan}
   \tdm\bbI \ \preceq\ \hat{ \bbH}(\bbw,{\tbtheta}) \ \preceq \ \tdM \bbI.
\end{equation} \end{assumption}
%
%
\begin{assumption}\normalfont\label{ass_bounded_stochastic_gradient_norm} There exists a constant $S^2$ such that for all variables $\bbw$ the second moment of the norm of the stochastic gradient satisfies
\begin{equation}\label{ekhtelaf}
   \mbE_{\bbtheta} \big{[} \| \hbs(\bbw_{t},\tbtheta_{t})\|^{2} \big{]} \leq S^{2}, 
\end{equation} \end{assumption}

%
\begin{assumption}\normalfont\label{ass_delta_less_m} The regularization constant $\delta$ is smaller than the smallest Hessian eigenvalue $\tdm$, i.e., $\delta<\tdm$.
\end{assumption}

Recall that according to Lemma \ref{flen} the update in \eqref{akbar} is a solution to \eqref{jadid} as long as the inner product $(\hbr_t-\delta\bbv_t)^T\bbv_t = \tbr_{t}^T\bbv_t>0$ is positive. Our first result is to show that selecting $\delta < \tdm$ as required by Assumption \ref{ass_delta_less_m} guarantees that this inequality is satisfied for all times $t$.

\begin{figure}
\centering
\includegraphics[width=\linewidth,height=0.48\linewidth]{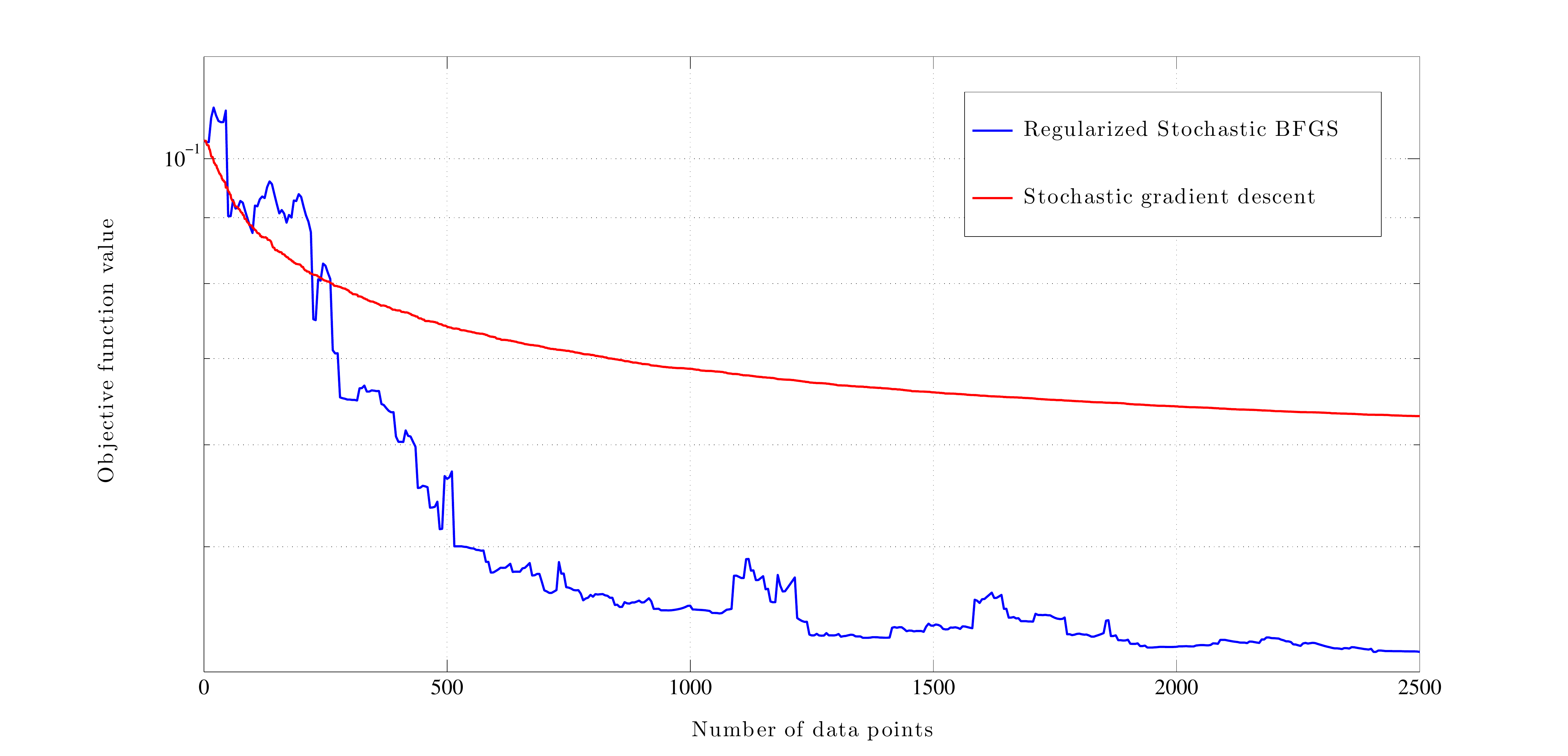}
\caption{Convergence of stochastic gradient descent and regularized stochastic BFGS for feature vectors of dimension $n=4$. Convergence of stochastic BFGS is faster than convergence of stochastic gradient descent (sample size $L=5$; stepsizes $\epsilon_t=\epsilon_{0}\tau / (\tau+t)$ with $\epsilon_{0} = 3\times10^{-2}$ and $\tau=10^2$; stochastic BFGS parameters  $\delta=10^{-3}$ and $\Gamma=10^{-4}$).}
\label{fig:4}
\end{figure}

%
\begin{lemma}\label{lecce}
Consider the modified stochastic gradient variation $\tbr_{t}$ defined in \eqref{chris2} and the variable variation $\bbv_{t}$ defined in \eqref{chris}. If assumptions \ref{ass_intantaneous_hessian_bounds} and \ \ref{ass_delta_less_m} are true, then, for all times $t$ it holds
\begin{equation}\label{claim23}
   \tbr_{t}^{T}\bbv_{t} \ =\ (\hbr_{t}-\delta \bbv_{t})^T\bbv_{t}\ \geq (\tdm-\delta) \|\bbv_{t}\|^{2} \ >\ 0.
\end{equation}\end{lemma}

%
The result in Lemma \ref{lecce} guarantees that the regularized stochastic BFGS algorithm as defined by recursive application of \eqref{eqn_sbfgs_dual_iteration}, \eqref{chris}, \eqref{chris2}, and \eqref{akbar} results in matrices $\hbB_t$ that solve \eqref{jadid}. In particular, this implies that $\hbB_t$ is positive definite with smallest eigenvalue not smaller than $\delta$, i.e., $\hbB_t\preceq\delta\bbI$. This implies that all the eigenvalues of $\hbB_{t}^{-1}$ are between $0$ and $1/\delta$ and that, as a consequence, the matrix $\hbB_t^{-1}+\Gamma \bbI$ is such that
\begin{equation}
   \Gamma \bbI\ \preceq\ \hbB_t^{-1}+\Gamma \bbI  \preceq\ (\Gamma+\frac{1}{\delta})\ \! \bbI.
\end{equation} 
Having matrices $\hbB_t^{-1}+\Gamma \bbI$ that are strictly positive definite with eigenvalues uniformly upper bounded by $\Gamma+(1/\delta)$ leads to the conclusion that if $\hbs(\bbw_{t},\tbtheta_{t})$ is a descent direction, the same holds true of $(\hbB_t^{-1}+\Gamma \bbI)\ \! \hbs(\bbw_{t},\tbtheta_{t})$. The stochastic gradient $\hbs(\bbw_{t},\tbtheta_{t})$ is not a descent direction in general, but we know that this is true for its conditional expectation $\mbE[\hbs(\bbw_{t},\tbtheta_{t}) \given \bbw_{t}] = \nabla_{\bbw} F(\bbw_{t})$. Therefore, we conclude that $(\hbB_t^{-1}+\Gamma \bbI)\ \! \hbs(\bbw_{t},\tbtheta_{t})$ is an average descent direction because $\mbE[(\hbB_t^{-1}+\Gamma \bbI)\ \!\hbs(\bbw_{t},\tbtheta_{t}) \given \bbw_{t}] = (\hbB_t^{-1}+\Gamma \bbI)\ \! \nabla_{\bbw} F(\bbw_{t})$. Having a displacement $\bbw_{t+1}-\bbw_t$ that is a descent direction on average implies convergence towards optimal arguments as we claim in the following theorem.

\begin{figure}
\centering
\includegraphics[width=\linewidth,height=0.48\linewidth]{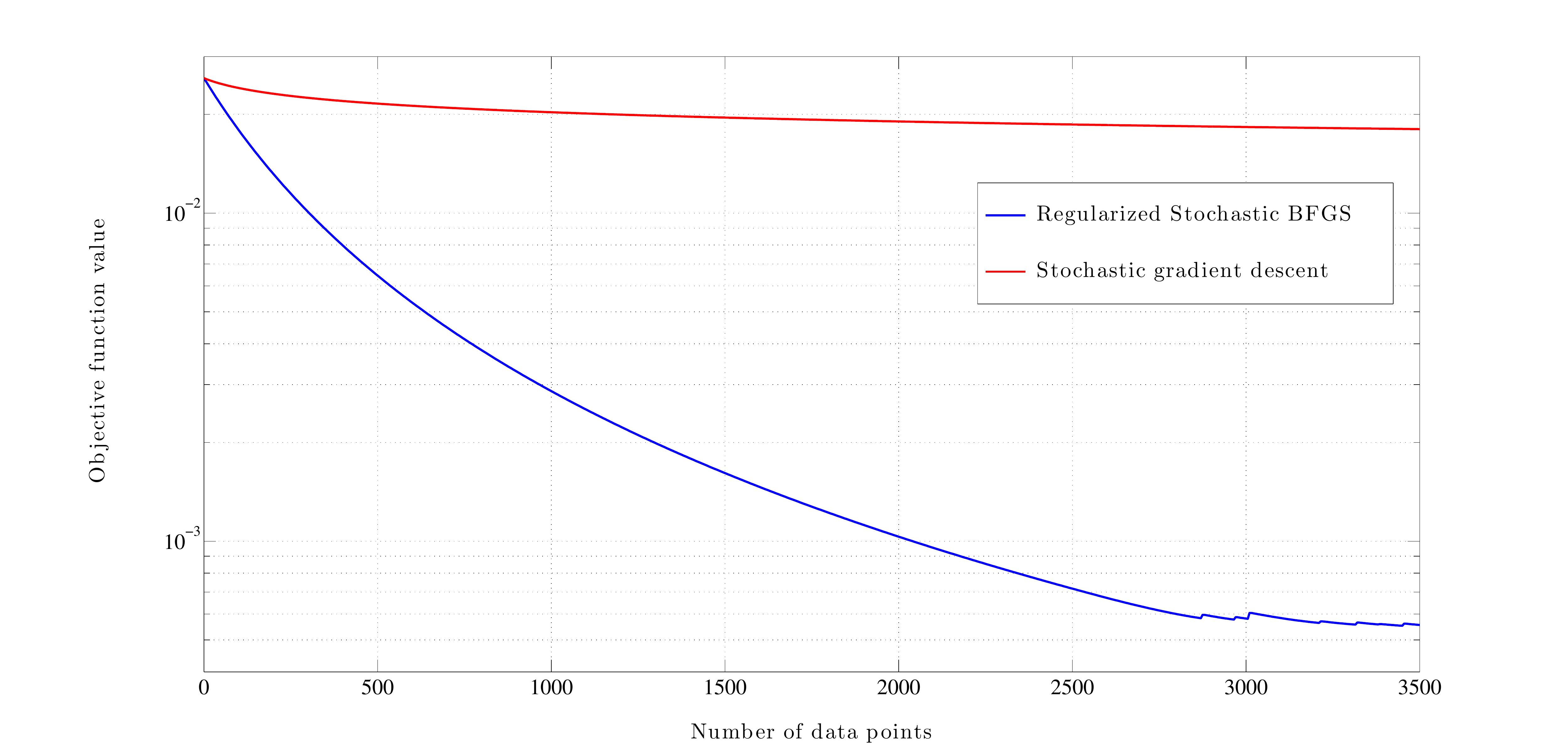}
\caption{Convergence of stochastic gradient descent and regularized stochastic BFGS for feature vectors of dimension $n=40$. Stochastic BFGS is still practicable whereas stochastic gradient descent becomes too slow for practical use (parameters as in Fig. \ref{fig:4}).}
\label{fig:5}
\end{figure}

%
\begin{thm}\label{convg}
Consider the regularized stochastic BFGS algorithm as defined by \eqref{eqn_sbfgs_dual_iteration}, \eqref{chris}, \eqref{chris2}, and \eqref{akbar}. If assumptions \ref{ass_intantaneous_hessian_bounds}-\ref{ass_delta_less_m} hold true and the sequence of stepsizes satisfies is nonsummable but square summable, i.e., if $   \sum_{t=0}^{\infty} \eps_t = \infty,$ and $\sum_{t=0}^{\infty} \eps_t^2 < \infty$, the limit infimum of the squared Euclidean distance to optimality $\| \bbw_{t}-\bbw^{*} \|^{2}$ satisfies
\begin{equation}\label{eqn_convg}
   \liminf_{t \to \infty }\| \bbw_{t}-\bbw^{*} \|^{2} = 0 \qquad \text{a.s.}
\end{equation} 
over realizations of the random samples $\{\tbtheta_t\}_{t=1}^\infty$.\end{thm}

Theorem \ref{convg} establishes convergence of the stochastic regularized BFGS algorithm summarized in Algorithm 1. In the proof of this result the lower bound in the eigenvalues of $\hbB_t$ enforced by the regularization in \eqref{akbar} plays a fundamental role. Roughly speaking, the lower bound in the eigenvalues of $\hbB_t$ results in an upper bound on the eigenvalues of $\hbB_t^{-1}$ which limits the effect of random variations on the stochastic gradient $\hbs(\bbw_{t},\tbtheta_{t})$. If this regularization is not implemented, i.e., if we keep $\delta=0$, we may observe catastrophic amplification of random variations of the stochastic gradient. This effect is indeed observed in the numerical experiments in Section \ref{sec:SVM}. The addition of the identity matrix bias $\Gamma\bbI$ in \eqref{eqn_sbfgs_dual_iteration} is also instrumental in the proof of Theorem \ref{convg}. This bias limits the effects of randomness in the curvature estimate $\hbB_t$. If random variations in the curvature estimate $\hbB_t$ result in a matrix $\hbB_t^{-1}$ with small eigenvalues the term $\Gamma\bbI$ dominates and \eqref{eqn_sbfgs_dual_iteration} reduces to stochastic gradient descent. This ensures continued progress towards the optimal argument $\bbw^*$. 

The convergence claim in Theorem \ref{convg}is complemented by a expected convergence rate result which we state in the following theorem.


\begin{thm}\label{theo_convergence_rate}
Consider the regularized stochastic BFGS algorithm as defined by \eqref{eqn_sbfgs_dual_iteration}-\eqref{akbar} and let the sequence of stepsizes be given by $\epsilon_{t}=\epsilon_{0}\tau / (\tau+t)$ with the parameter $\eps_0$ sufficiently small and the parameter $\tau$ sufficiently large so as to satisfy the inequality
\begin{equation}\label{eqn_thm_cvg_rate_10}
    2\  \epsilon_{0} \tau \Gamma >1\ .
\end{equation} 
If assumptions \ref{ass_intantaneous_hessian_bounds} and \ref{ass_bounded_stochastic_gradient_norm} hold true the difference between the expected objective value $\E {F(\bbw_{t})}$ at time $t$ and the optimal objective $F(\bbw^*)$ satisfies
\begin{equation}\label{eqn_thm_cvg_rate_20}
\E {F(\bbw_{t})}- F(\bbw^*)\ \leq\ \frac{\xi}{\tau+t}\ ,
\end{equation}
where the constant $\xi$ satisfies
\begin{equation}\label{eqn_thm_cvg_rate_30}
 \xi\ =\ \max\ \left\{ \frac{\epsilon_{0}^{2}\   \tau^{2} K}{2 \epsilon_{0} \tau \Gamma -1}\ , (1+\tau)  (F(\bbw_{0}) -\ F(\bbw^*))  \right\} .
\end{equation}
\end{thm}

%

Theorem \ref{theo_convergence_rate} shows the convergence rate of regularized stochastic BFGS is at least linear in terms of the expectation of the objective function. This rate is typical of stochastic optimization algorithms and, in that sense, no better than stochastic gradient descent. While the convergence rate doesn't change, improvements in convergence time are marked as we illustrate with the numerical experiments of the following section.


\vspace{-1mm}

\section{Numerical Analysis}
\label{sec:SVM}

\begin{figure}[t]
\centering
\includegraphics[width=\linewidth,height=0.48\linewidth]{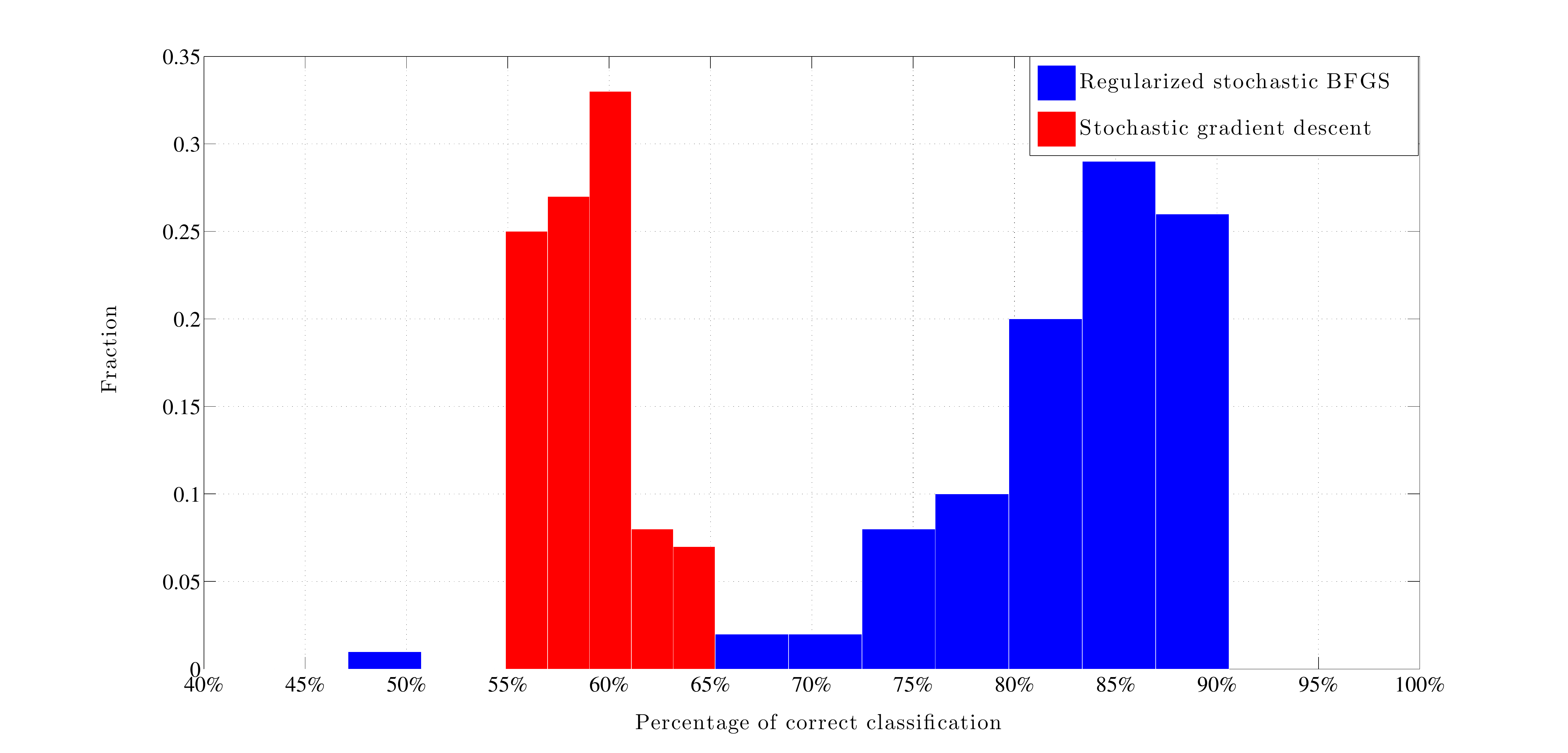}
\caption{Histogram of correct classification percentages for dimension $n=4$ and training set of size $N=2.5\times 10^3$. Vectors computed by stochastic BFGS outperform those computed via stochastic gradients and are not far from the accuracy of clairvoyant classifiers (test sets contain $10^4$ samples; histogram is across $10^3$ realizations; parameters as in Fig. \ref{fig:4}).}
\label{accuracy}
\end{figure}

We test Algorithm 1 when using the squared hinge loss $l((\bbx,y);\bbw)=\max(0,1-y(\bbx^{T}\bbw))^{2}$ in \eqref{SVM}. The training set $\ccalS = \{ (\bbx_{i},y_{i}) \}_{i=1}^{N}$ contains $N=10^4$ feature vectors half of which belong to the class $y_i=-1$ with the other half belonging to the class $y_i=1$. For the class $y_i=-1$ each of the $n$ components of each of the feature vectors $\bbx_i\in\reals^n$ is chosen uniformly at random from the interval $[-0.8,0.2]$. Likewise, each of the $n$ components of each of the feature vectors $\bbx_i\in\reals^n$ is chosen uniformly at random from the interval $[-0.2,0.8]$ for the class $y_i=1$. The overlap in the range of the feature vectors is such that the classification accuracy expected from a clairvoyant classifier that knows the statistic model of the data set is less than $100\%$. Exact values can be computed from the Irwin-Hall distribution \cite{Johnson}. For $n=4$ this amounts to $98\%$.

We set the parameter $\lambda$ in \eqref{SVM} to $\lambda=10^{-3}$. Since the Hessian eigenvalues of $f(\bbw,\bbtheta) :=\lambda\|\bbw\|^2/2 + l((\bbx_{i},y_{i});\bbw)$ are, at least, equal to $\lambda$ this implies that the eigenvalue lower bound $\tdm$ is such that $\tdm\geq\lambda=10^{-3}$. Thus, we set the BFGS regularization parameter to $\delta=\lambda=10^{-3}$. Further set the minimum progress parameter in \eqref{stochastic_gradient} to $\Gamma=10^{-4}$ and the sample size for computation of stochastic gradients to $L=5$. Stepsizes are of the form $\epsilon_t=\epsilon_{0}\tau / (\tau+t)$ with $\epsilon_{0} = 3\times10^{-2}$ and $\tau=10^2$. We compare the behavior of stochastic gradient descent and stochastic BFGS for a small dimensional problem with $n=4$ and a large problem with $n=40$. For stochastic gradient descent the sample size in \eqref{stochastic_gradient} is $L=1$ and we use the same stepsize sequence used for stochastic BFGS.

An illustration of the relative performances of stochastic gradient descent and BFGS for $n\!=4$ is presented in Fig. \ref{fig:4}. The value of the objective function $F(\bbw_t)$ is represented with respect to the number of feature vectors processed, which is given by the product $Lt$ between the iteration index and the sample size used to compute stochastic gradients. This is done because the sample sizes in stochastic BFGS ($L=5$) and stochastic gradient descent ($L=1$) are different. The curvature correction of stochastic BFGS results in significant reductions in convergence time. E.g., Stochastic BFGS achieves an objective value of $F(\bbw_t)=6.5\times10^{-2}$ upon processing of $Lt=315$ feature vectors. To achieve the same objective value $F(\bbw_t)=6.5\times10^{-2}$ stochastic gradient descent processes {$1.74\times10^3$} feature vectors. Conversely, after processing $Lt=2.5\times10^3$ feature vectors the objective values achieved by stochastic BFGS and gradient descent  are $F(\bbw_t) =4.14\times10^{-2}$ and $F(\bbw_t)=6.31\times10^{-2}$, respectively. 

The performance difference between the two methods is larger for feature vectors of larger dimension $n$. The plot of the value of the objective function $F(\bbw_t)$ with respect to the number of feature vectors processed $Lt$ is shown in Fig. \ref{fig:5} for $n=40$. The convergence time of stochastic BFGS increases but is still acceptable. For stochastic gradient descent the algorithm becomes unworkable. After processing $3.5\times 10^{3}$ stochastic BFGS reduces the objective value to  $F(\bbw_t)=5.55\times10^{-4}$ while stochastic gradient descent has barely made progress at $F(\bbw_t)=1.80\times10^{-2}$.

Differences in convergence times translate into differences in classification accuracy when we process all $N$ vectors in the training set. This is shown for dimension $n=4$ and training set size $N=2.5\times10^3$ in Fig. \ref{accuracy}. To build Fig. \ref{accuracy} we process $N=2.5\times10^3$ feature vectors with stochastic BFGS and stochastic gradient descent with the same parameters used in Fig. \ref{fig:4}. We then use these vectors to classify $10^4$ observations in the test set and record the percentage of samples that are correctly classified. The process is repeated $10^3$ times to estimate the probability distribution of the correct classification percentage represented by the histograms shown. The dominance of stochastic BFGS with respect to stochastic gradient descent is almost uniform. The vector $\bbw_t$ computed by stochastic gradient descent classifies correctly at most $65\%$ of the of the feature vectors in the test set. The vector $\bbw_t$ computed by stochastic BFGS exceeds this accuracy with probability $0.98$. Perhaps more relevant, the classifier computed by stochastic BFGS achieves a mean classification accuracy of $82.2\%$ which is not far from the clairvoyant classification accuracy of $98\%$. Although performance is markedly better in general, stochastic BFGS fails to compute a working classifier with probability $0.02$.

\begin{figure}[t]
\centering
\includegraphics[width=\linewidth,height=0.48\linewidth]{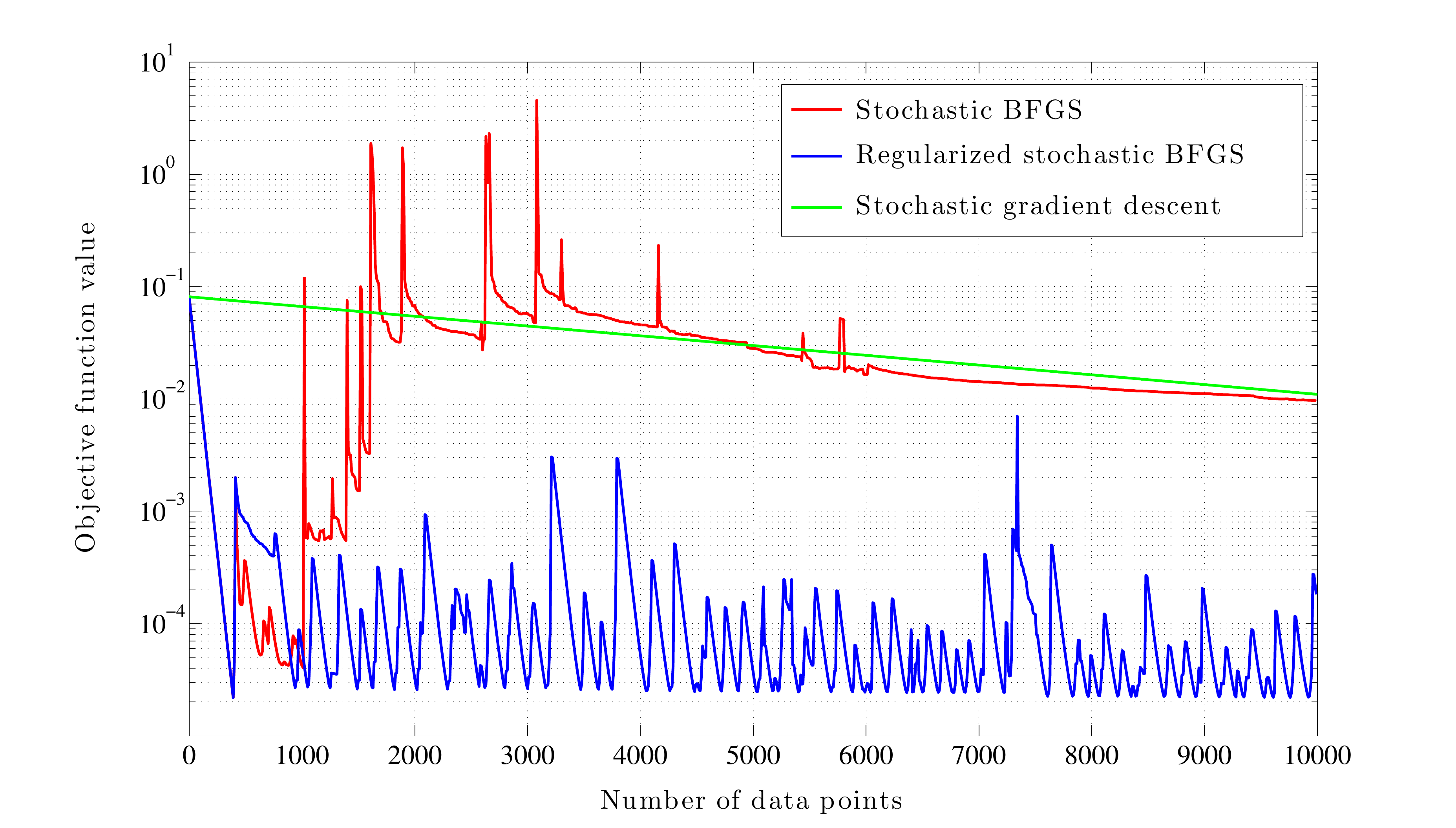}
\caption{Comparison of gradient descent, regularized stochastic BFGS, and (non regularized) stochastic BFGS. The regularization is fundamental to control the erratic behavior of stochastic BFGS (sample size $L=5$; constant stepsize $\epsilon_{t} = 10^{-1}$; stochastic BFGS parameters  $\delta=10^{-3}$ and $\Gamma=10^{-4}$, feature vector dimension $n=10$).}
\label{jumps}
\end{figure}

We also investigate the difference between regularized and non-regularized versions of stochastic BFGS for feature vectors of dimension $n=10$. Observe that non-regularized stochastic BFGS corresponds to making $\delta=0$ and $\Gamma=0$ in Algorithm 1. To illustrate the advantage of the regularization induced by the proximity requirement in \eqref{jadid}, as opposed to the non regularized proximity requirement in \eqref{jadid_prelim}, we keep a constant stepsize $\epsilon_{t}=10^{-1}$. The corresponding evolutions of the objective function values $F(\bbw_t)$ with respect to the number of feature vectors processed $Lt$ are shown in Fig. \ref{jumps} along with the values associated with stochastic gradient descent. As we reach convergence the likelihood of having small eigenvalues appearing in $\hbB_t$ becomes significant. In regularized stochastic BFGS this results in recurrent jumps away from the optimal classifier $\bbw^*$. However, the regularization term limits the size of the jumps and further permits the algorithm to consistently recover a reasonable curvature estimate. In Fig. \ref{jumps} we process $10^4$ feature vectors and observe many occurrences of small eigenvalues. However, the algorithm always recovers and heads back to a good approximation of $\bbw^*$. In the absence of regularization small eigenvalues in $\hbB_t$ result in larger jumps away from $\bbw^*$. This not only sets back the algorithm by a much larger amount than in the regularized case but also results in a catastrophic deterioration of the curvature approximation matrix $\hbB_t$. In Fig. \ref{jumps} we observe recovery after the first two occurrences of small eigenvalues but eventually there is a catastrophic deviation after which non-regularized stochastic BFSG behaves not better than stochastic gradient descent.

\section{Conclusions}

We considered the problem of determining the separating hyperplane of a support vector machine using stochastic optimization. In order to handle large scale problems with reasonable convergence times we adapted a regularized stochastic version of the Broyden, Fletcher, Goldfarb, and Shanno (BFGS) quasi-Newton method\cite{cMokhtariRibeiro13}. We derived theoretical convergence guarantees that are customary of stochastic optimization and illustrated improvements in convergence time through numerical analysis. 



\bibliographystyle{IEEEtran}
  \bibliography{bmc_article,01_my_journals_2005_2008,bib_files/02_my_conferences_2009_2010,bib_files/02_my_conferences_2004_2008,bib_files/01_my_journals_2005_2008,bib_files/01_my_journals_2009_2010,bib_files/02_my_conferences_2011,bib_files/02_my_conferences_2012,bib_files/01_my_journals_2012,bib_files/01_my_journals_2011,bib_files/04_bib_cross_layer,bib_files/yichuan_1,bib_files/yichuan_2}
   \end{document}